\documentclass[letterpaper, 10pt, conference]{ieeeconf}   % IEEE 会议模板
\IEEEoverridecommandlockouts                               % 解锁 \thanks
\usepackage{float}
\usepackage{graphicx}
\usepackage{booktabs}
\usepackage{multirow}
\usepackage{xcolor}
\usepackage{amsmath}
\usepackage{algorithm, algorithmicx}
\usepackage{fancyvrb}
\usepackage{tikz}
\usepackage{graphicx}
\usepackage{hyperref}

\usepackage{algorithmicx}
\usepackage{algpseudocode}
\title{UnderwaterVLA: Dual-brain Vision-Language-Action architecture for Autonomous Underwater Navigation}
\author{Zhangyuan Wang$^{1}$$^{2}$ \textsuperscript{\#}, Yunpeng Zhu$^{1}$$^{3}$ \textsuperscript{\#}, Yuqi Yan$^{1}$, Xiaoyuan Tian$^{1}$, Xinhao Shao$^{1}$,\\Meixuan Li$^{4}$, Weikun Li$^{1}$, Guangsheng Su$^{1}$, Weicheng Cui$^{1}$, Dixia Fan$^{1}$*%
\thanks{*Corresponding author: Dixia Fan}% <-this % stops a space
\thanks{E-mails: fandixia@westlake.edu.cn}%
\thanks{\textsuperscript{\#}These authors contributed equally to this work.}
\thanks{$^{1}$School of Engineering, Westlake University, Hangzhou, 310030, China.}%
\thanks{$^{2}$College of Information Science \& Electronic Engineering, Zhejiang University, Hangzhou, 310027, China.}
\thanks{$^{3}$College of Environmental and Resource Sciences, Zhejiang University, Hangzhou, 310027, China.}
\thanks{$^{4}$Australian National University, Canberra, ACT 2601, Australia.}
\thanks{\textbf{Code:} \url{https://anonymous.4open.science/r/UnderwaterVLA-8FC8/ }}
}

\begin{document}

\maketitle
\thispagestyle{empty}
\pagestyle{empty}

%---------- 摘要 ----------
\begin{abstract}
This paper presents UnderwaterVLA, a novel framework for autonomous underwater navigation that integrates multimodal foundation models with embodied intelligence systems. Underwater operations remain difficult due to hydrodynamic disturbances, limited communication bandwidth, and degraded sensing in turbid waters. To address these challenges, we introduce three innovations. First, a dual-brain architecture decouples high-level mission reasoning from low-level reactive control, enabling robust operation under communication and computational constraints. Second, we apply Vision–Language–Action (VLA) models to underwater robotics for the first time, incorporating structured chain-of-thought reasoning for interpretable decision-making. Third, a hydrodynamics-informed Model Predictive Control (MPC) scheme compensates for fluid effects in real time without costly task-specific training. Experimental results in field tests show that UnderwaterVLA reduces navigation errors in degraded visual conditions while maintaining higher task completion by 19$\%$ to 27\% over baseline. By minimizing reliance on underwater-specific training data and improving adaptability across environments, UnderwaterVLA provides a scalable and cost-effective path toward the next generation of intelligent AUVs.
\end{abstract}

%---------- IEEE 关键词 ----------

% \begin{IEEEkeywords}

%---------- 正文 ----------
\section{Introduction}
\label{sec:introduction}

Autonomous Underwater Vehicles (AUVs) are indispensable tools for marine science and industry, supporting missions ranging from seabed mapping and ecological monitoring to subsea infrastructure inspection and resource exploration \cite{bovio2006autonomous}. Their ability to operate untethered in extreme environments provides unique advantages over remotely operated vehicles (ROVs), including extended range, reduced human supervision, and lower operational costs \cite{leonard2016autonomous}. AUVs have enabled breakthroughs in deep-sea research and offshore engineering by delivering high-resolution data from environments that are otherwise inaccessible.

Yet, AUV autonomy remains fundamentally constrained by the hostile and unpredictable nature of the ocean. Unlike aerial or terrestrial domains, underwater environments present a combination of challenges:

\begin{enumerate}
    \item \textbf{Strong, nonlinear hydrodynamics}, where ocean currents, turbulence, and vortical structures introduce unmodeled disturbances that degrade stability and control \cite{fossen2011handbook,antonelli2014underwater}.
    \item \textbf{Extreme physical conditions}, including high hydrostatic pressures, spatially varying salinity, and thermocline effects, which alter vehicle buoyancy and structural reliability \cite{cong2021underwater}.
    \item \textbf{Severe perception difficulties}, such as turbidity, backscatter, and non-uniform illumination, which compromise optical sensors; meanwhile, sonar imaging suffers from low resolution and multipath interference \cite{zhang2023autonomous, mcconnell2022perception}.
    \item \textbf{Communication constraints}, where radio is ineffective underwater and acoustic channels are bandwidth-limited, delay-prone, and vulnerable to noise \cite{stojanovic2008underwater}.
\end{enumerate}

Conventional approaches—including proportional–integral–derivative (PID) controllers, adaptive model-based control, and deep reinforcement learning—have shown partial success, but often lack generalization and robustness in unstructured marine environments \cite{lu2023reinforcement, qiao2023survey}. Recent efforts such as factor-graph-based cooperative localization \cite{velasco2025factor} and data-informed domain randomization for reinforcement learning \cite{biao2025research, fan2025long} have improved resilience but remain constrained by reliance on accurate modeling and costly task-specific training. This highlights the need for autonomy frameworks capable of generalizing across tasks and adapting to uncertainty without prohibitive data requirements.

The emerging Vision–Language–Action (VLA) paradigm offers such potential. By unifying perception, natural language understanding, and action generation, VLA systems enable multimodal reasoning and robust task generalization. These models have achieved remarkable success in diverse domains, including industrial manipulation, autonomous driving, aerial robotics, and quadruped locomotion \cite{lynch2023interactive, 10906781, ding2024quar,sautenkov2025uav}. For instance, OpenVLA demonstrates unified multimodal control and instruction following \cite{11153233}, while DiffusionVLA shows how large-scale diffusion and autoregressive models can enhance general-purpose robot performance \cite{wen2025diffusionvla}. QUAR-VLA has successfully applied VLA models to quadrupeds for locomotion and manipulation \cite{ding2024quar}. Such advances indicate strong potential for VLA to move beyond task-specific controllers toward general-purpose autonomy with natural language interaction and contextual understanding.

However, VLA remains largely unexplored in underwater robotics \cite{white2024new}. Three barriers explain this gap: (i) \textbf{No underwater-suited hierarchy}: Existing end-to-end VLA models cannot effectively decouple high-level mission planning from low-level real-time control under the stringent communication and computational constraints of underwater operations, leading to insufficient system robustness.
(ii) \textbf{Data-hungry policies}: Conventional VLA require vast amounts of expensive underwater demonstration data to learn policies, severely limiting scalability and generalization due to prohibitive data collection costs \cite{biao2025research}.
(iii) \textbf{No real-time hydro-compensation}: Nonlinear fluid dynamics (e.g., turbulence, vortices) significantly disrupt AUV motion; existing VLA or learning-based controllers lack embedded physical priors and cannot perform real-time dynamic compensation without additional task-specific training, resulting in degraded navigation accuracy \cite{elmezain2025advancing, cong2021underwater}. 
 These challenges have previously hindered the application of VLA frameworks to AUVs, despite their transformative promise. Importantly, recent advances in VLA research increasingly converge toward model-based, zero-data training paradigms — leveraging physical priors \cite{Yin2025GCVLN}.

To address these issues, we present Underwater VLA, the first VLA framework specifically designed for AUVs. Our system systematically addresses the unique challenges of underwater environments by introducing a zero-data training methodology, cross-domain adaptation with physical priors, and a dual-brain architecture that integrates high-level reasoning with low-level control for robust autonomy.

The main contributions of this paper are summarized as follows:
\begin{itemize}
\item \textbf{Underwater VLA framework}: We introduce the first VLA framework tailored for AUVs, systematically addressing underwater-specific challenges such as limited communication, harsh sensing conditions, and costly data collection.
\item \textbf{Zero-data training methodology}: We propose a training strategy that minimizes reliance on underwater demonstrations by leveraging pre-trained multimodal foundation models with underwater-specific transfer learning.
\item \textbf{Dual-brain architecture}: We design a hybrid control structure that decouples high-level deliberative reasoning from low-level reactive control, ensuring reliable autonomy under computational constraints and enhancing safety.
\item \textbf{Experimental validation}: We demonstrate that Underwater VLA achieves superior language understanding, task generalization, and mission success rates compared to conventional control and previous VLA baseline.
\end{itemize}
%conventional control and reinforcement learning baselines

The remainder of this paper is organized as follows: Section I reviews related work on AUV autonomy and VLA frameworks. Section II details the methodology of Underwater VLA. Section III presents experimental results and discussion. Section IV concludes the paper with future directions.

\section{Methodology}
\subsection{Dual-Brain Architecture}
Our dual-brain model framework addresses underwater communication constraints by decoupling strategic planning from real-time execution:
\begin{equation}
\Phi_{\text{cloud}} \xrightarrow{\text{QVQmax}} \{\mathcal{S}_1, \mathcal{S}_2, \dots, \mathcal{S}_N\}, \quad \Phi_{\text{local}} \xrightarrow{\text{Qwen-VL}} \mathcal{J}_t
\end{equation}
where $\{\mathcal{S}_1, \dots, \mathcal{S}_N\}$ denotes a sequence of high-level sub-goals (steps) generated by the cloud brain for long-horizon mission decomposition, and $\mathcal{J}_t$ represents the local JSON-formatted control decision at time $t$. 

The \textbf{cloud brain} operates intermittently during surfacing events, producing coarse-grained, long-horizon plans that are decomposed into executable sub-tasks. The \textbf{cerebellum} (on-device brain) then assumes control, executing these sub-goals through step-by-step closed-loop perception--action cycles, optimized for real-time operation under bandwidth-limited conditions.

\begin{figure*}[!htbp]
  \centering
  \includegraphics[width=0.9\textwidth]{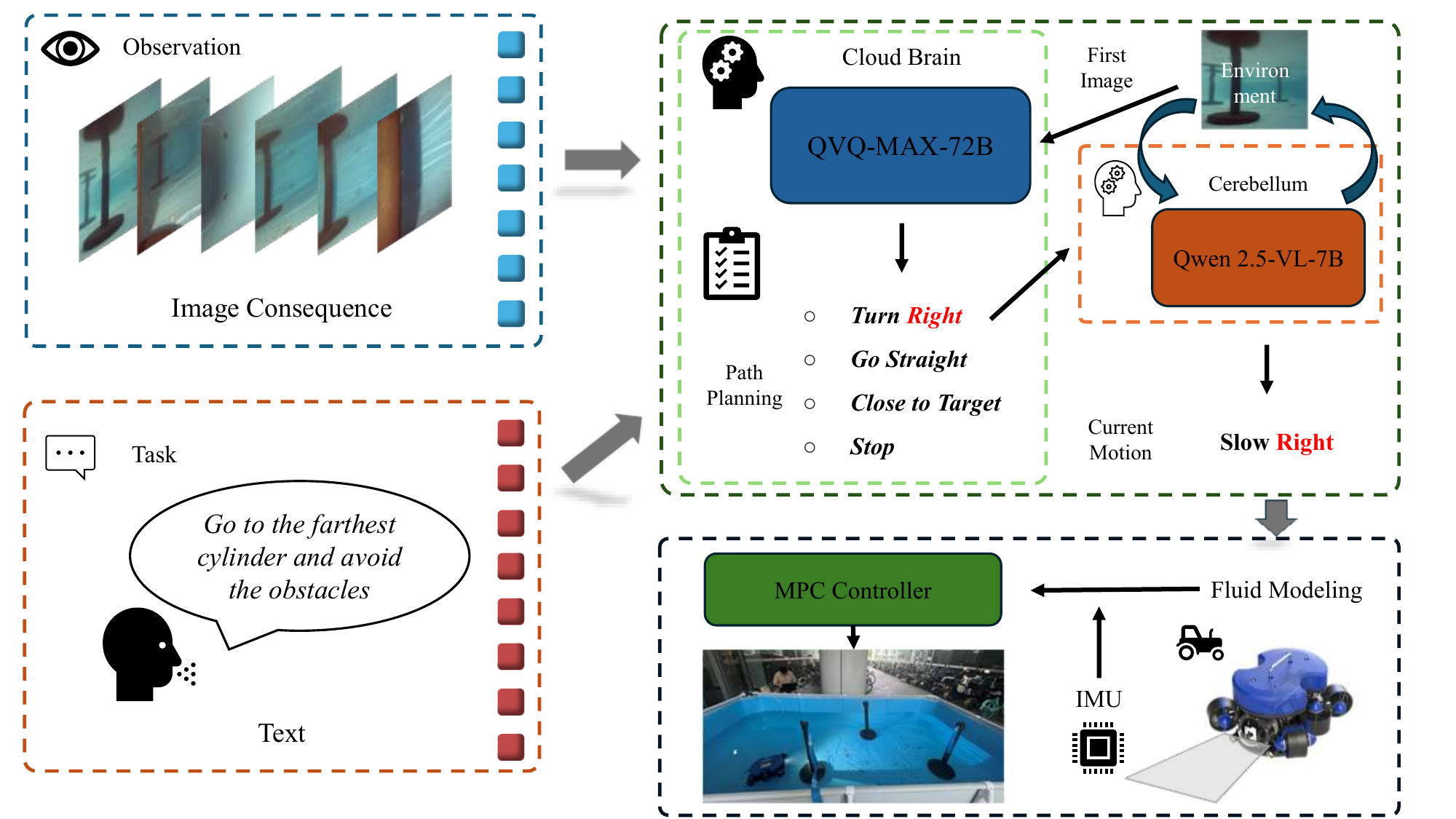} % Replace with the actual filename of your image
  \caption{
    Schematic of the Dual-Brain Cooperative Framework: Long-horizon planning (cloud) is decomposed into sequential sub-goals, while local execution (cerebellum) performs closed-loop, step-wise control via online/offline VLMs.}
  \label{fig:Model_Schematic}
\end{figure*}

\subsection{Prompt Engineering Framework}
\label{sssec:cot}

To enhance decision interpretability and facilitate failure diagnosis, we enforce explicit Chain-of-Thought (CoT) reasoning in both the \textbf{cloud brain} and the \textbf{cerebellum}. Each module outputs structured rationales alongside control decisions, ensuring traceability of intent and action.

The \textbf{cloud brain} decomposes high-level instructions into temporally ordered sub-tasks via CoT prompting. For example, given ``Navigate to the coral reef while avoiding shipwrecks,'' it generates:
\begin{enumerate}
\item \texttt{Sub-task 1}: Localize via sonar landmarks.
\item \texttt{Sub-task 2}: Plan collision-free coarse path.
\item \texttt{Sub-task 3}: Transmit waypoint A with medium velocity.
\end{enumerate}

The \textbf{cerebellum} executes each sub-task under bandwidth constraints, outputting a JSON-structured control tuple at each timestep:
\begin{verbatim}
{
  "reasoning": "Obstacle detected left; 
  turning right.",
  "decision": "right",
  "velocity": "medium",
  "sub_task_done": false,
  "mission_done": false
}
\end{verbatim}

This schema enforces mandatory fields for reasoning, action, velocity, and task status, ensuring consistent logging and enabling real-time monitoring of decision processes.

\subsection{Zero-Data Model Predictive Control for Discrete Motion Execution}  
\label{ssec:mpc_motion}

\textbf{Time-Constrained Motion Profile with Dual Motion Modes:}
Given a discrete VLA command $\{\mathcal{D}, \mathcal{V}\}$ specifying direction and velocity level, the model predictive controller generates two distinct motion profiles with strict timing constraints. 

For \emph{translational motion} commands (forward/backward), the velocity profile follows:
\begin{equation}
v(t) = 
\begin{cases} 
    v_{\text{cmd}} \cdot \min(5t, 1) & 0 \leq t < 0.2 \\
    v_{\text{cmd}} & 0.2 \leq t < 0.5 \\
    v_{\text{cmd}} \cdot \max(1 - 2(t-0.5), 0) & 0.5 \leq t \leq 1
\end{cases}
\end{equation}
where $v_{\text{cmd}}$ represents the target velocity magnitude corresponding to the discrete velocity level (0.2 m/s for low, 0.5 m/s for medium, 0.8 m/s for high). 

For \emph{rotational motion} commands (turn left/turn right), the angular velocity profile follows:
\begin{equation}
r(t) = 
\begin{cases} 
    r_{\text{cmd}} \cdot \min(5t, 1) & 0 \leq t < 0.2 \\
    r_{\text{cmd}} & 0.2 \leq t < 0.5 \\
    r_{\text{cmd}} \cdot \max(1 - 2(t-0.5), 0) & 0.5 \leq t \leq 1
\end{cases}
\end{equation}
where $r_{\text{cmd}}$ represents the target angular velocity magnitude (0.5 rad/s for low, 1.0 rad/s for medium, 1.5 rad/s for high). The complete motion execution cycle is strictly constrained to 1 second duration for both modes.

\textbf{Hydrodynamic-Informed MPC Formulation with Mode Switching:}
The controller optimizes thrust commands $\boldsymbol{\tau}$ (control input vector) to track the reference profile while compensating for fluid dynamic effects through a mode-dependent cost function:
\begin{figure*}[h   ]
\centering
\begin{equation}
\min_{\boldsymbol{\tau}} \int_{0}^{1} \left\{
\begin{array}{l}
\beta\left\|v(t)-v_{\text{ref}}(t)\right\|^{2} + \gamma\left\|\boldsymbol{\tau}(t)\right\|^{2} + \delta\left\|F_{\text{drag},v}(v(t))\right\|^{2} \\
\beta\left\|\theta(t)-\theta_{\text{ref}}(t)\right\|^{2} + \gamma\left\|\boldsymbol{\tau}(t)\right\|^{2} + \delta\left\|\tau_{\text{drag},r}(r(t))\right\|^{2}
\end{array}
\right. dt
\label{eq:mpc_formulation}
\end{equation}
\end{figure*}
where $\beta$ weights the tracking error between actual velocity $v(t)$ and reference velocity $v_{\text{ref}}(t)$ for translational motion, or between actual angle $\theta(t)$ and reference angle $\theta_{\text{ref}}(t)$ for rotational motion. The term $\gamma$ penalizes control effort magnitude $\|\boldsymbol{\tau}(t)\|$, while $\delta$ weights the compensation for translational drag force $F_{\text{drag},v}$ and rotational drag torque $\tau_{\text{drag},r}$.

The quadratic drag forces are modeled as:
\begin{align}
F_{\text{drag},v} &= D_v v|v| \\
\tau_{\text{drag},r} &= D_r r|r|
\end{align}
where $D_v$ represents the translational drag coefficient (kg/m), and $D_r$ represents the rotational drag coefficient (kg·m²·s/rad²).

\textbf{Real-Time Fluid Adaptation with Parameter Estimation:}
Drag coefficients are estimated online during execution using inertial and dynamic measurements:
\begin{equation}
\hat{D}_v = \frac{\tau_v - M\dot{v}}{v|v|}, \quad \hat{D}_r = \frac{\tau_r - I_z\dot{r}}{r|r|}
\end{equation}
where $M$ denotes the system mass including added mass effects (kg), $I_z$ represents the moment of inertia about the vertical axis (kg·m²), $\dot{v}$ and $\dot{r}$ are acceleration measurements from IMU, and $\tau_v$, $\tau_r$ are the thrust components.

\textbf{MPC Execution with Mode Selection:}
\begin{algorithm}[H]
\caption{Multi-Mode Motion Execution with Fluid Compensation}
\begin{algorithmic}[1]
\Require VLA command $\{\mathcal{D}, \mathcal{V}\}$ with type specification
\If{$\mathcal{D} \in \{\texttt{forward}, \texttt{backward}\}$}
    \State Set motion mode to \texttt{translation}
    \State Set $v_{\text{cmd}}$ based on $\mathcal{V}$ level
    \State Define reference profile $v_{\text{ref}}(t)$ per Eq. (1)
\ElsIf{$\mathcal{D} \in \{\texttt{left}, \texttt{right}\}$}
    \State Set motion mode to \texttt{turning}
    \State Set $r_{\text{cmd}}$ based on $\mathcal{V}$ level
    \State Define reference profile $\theta_{\text{ref}}(t)$ per Eq. (2)
\EndIf
\For{$t = 0$ to $1$ s with control interval $\delta t = 0.02$ s}
    \State Measure current velocity $v_t$ or angular rate $r_t$
    \State Compute appropriate reference signal based on mode
    \State Estimate drag coefficients $\hat{D}_v$ or $\hat{D}_r$ via Eq. (4)
    \State Calculate drag forces/moments $F_{\text{drag}}$ or $\tau_{\text{drag},r}$
    \State Solve MPC optimization (Eq. 3) for optimal thrust $\boldsymbol{\tau}_t^*$
    \State Apply thruster command $\mathbf{u}_t = \mathbf{B}^\dagger \boldsymbol{\tau}_t^*$
\EndFor
\State Apply active braking until $\|v\| < 0.01$ m/s or $|r| < 0.01$ rad/s
\end{algorithmic}
\end{algorithm}

\textbf{Fluid–Structure Interaction with Mode-Specific Considerations:} The MPC formulation explicitly incorporates hydrodynamic effects across different motion phases. During the \textit{acceleration phase} (0–0.2,s), the controller compensates for added mass effects, modeled as $M_A = \rho \nabla C_A$, where $\rho$ is the water density (kg/m³), $\nabla$ is the displaced volume (m³), and $C_A$ is the added-mass coefficient. In the \textit{constant-velocity phase} (0.2–0.5,s), quadratic drag is addressed using $F_D = \tfrac{1}{2}\rho C_D A v^2$ for translation and $\tau_D = \tfrac{1}{2}\rho C_{Dr} L^4 r|r|$ for turning, where $C_D$ and $C_{Dr}$ denote translational and rotational drag coefficients, $A$ is the characteristic area (m²), $L$ is the characteristic length (m), and $r$ is the yaw rate. During \textit{deceleration} (0.5–1.0,s), the formulation leverages natural fluid damping to achieve smooth braking.

\textbf{Methodological Advantages:} The integrated dual-brain MPC approach provides several benefits for discrete underwater control. It ensures deterministic execution with exact 1-second motion cycles for both translation and rotation. By compensating in real time for phase-specific hydrodynamic effects, the controller reduces energy consumption and improves efficiency. The formulation further guarantees precise stopping with minimal residual velocity or angular drift. Importantly, the method operates in a \textit{zero-data regime}, relying on physics-based modeling rather than task-specific training, enabling robust adaptation to varying hydrodynamic conditions without costly data collection.

\section{Experiments}
\subsection{Case illustration: vision-guided navigation}

\begin{figure*}[!htbp]
  \centering
  \includegraphics[width=0.78\textwidth]{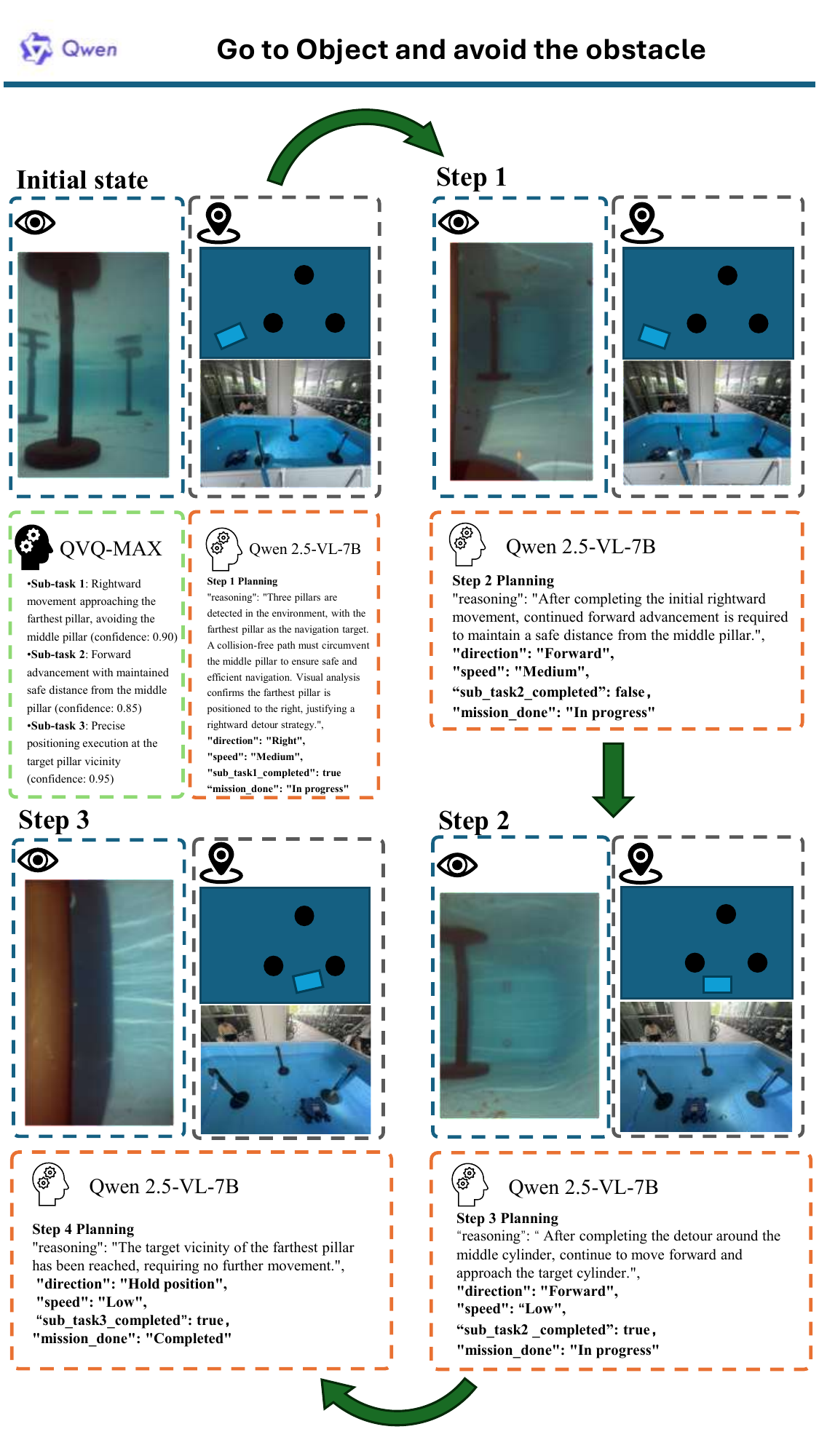} % Replace with the actual filename of your image
    \caption{Experimental demonstration of sequential navigation and obstacle avoidance in a cluttered underwater testbed. The AUV autonomously identifies the farthest target and executes collision-free maneuvers while maintaining robust trajectory tracking.
}
  \label{fig:Task_Study}
\end{figure*}

We conducted the experiment in a controlled laboratory tank containing three vertical cylindrical obstacles. The robot started from a predefined position and was tasked with reaching the right-most pillar—the farthest target—while avoiding collisions. At the initial state, the visual-perception module captured a single image of the scene and extracted the spatial arrangement of the three pillars. Using the QVQ-MAX model as the \textit{cloud  brain}, the system designated the farthest pillar as the navigation goal and the middle pillar as a potential obstacle. This high-level semantic description was transmitted to Qwen 2.5-VL-7B, which functioned as the \textit{cerebellum}. The cerebellum entered a standby mode, prepared for lateral maneuvering, while the robot completed environment modeling and goal identification.

Because the middle pillar lay directly ahead, a straight-line approach was deemed unsafe. In Step~1, the planner generated a detour strategy commanding a rightward motion. The cloud  brain issued a discrete “Right” directive based on the macro-policy “right is safe,” and the cerebellum translated it into a medium-speed clockwise turn. The robot laterally deviated toward the far pillar while maintaining clearance from the middle pillar, thereby completing the preliminary obstacle-avoidance maneuver.

After this deviation, the robot was positioned to the right of the middle pillar and needed to advance further to preserve clearance while closing the distance to the target. In Steps~2 and~3, the cloud  brain consecutively issued two “Forward” commands. In Step~2, the cerebellum drove the robot straight ahead at medium speed, ensuring steady progress while maintaining the safety margin. In Step~3, once the obstacle was fully cleared, the speed was reduced to “Low” for precise approach to the target pillar. The concatenation of these two forward motions produced a straight trajectory following the initial curved detour.

Finally, when visual feedback confirmed that the robot had reached the vicinity of the farthest pillar, the cloud brain transmitted a “Hold position” command. The cerebellum immediately halted the robot and maintained zero velocity. The task status was updated to “Completed,” concluding the obstacle-avoidance and goal-reaching sequence successfully coordinated by the dual-brain architecture.

\subsection{Quantitative performance evaluation against baseline methods}
RL and classical control are inherently task-specific and lack the capacity to interpret high-level, language-based instructions. 
Therefore, we compared our method against the baseline results  QUAR-VLA reported in \cite{ding2024quar}. While the baseline was evaluated in simulation, our experiments were conducted in more challenging real-world conditions. Even under these settings, the proposed hierarchical dual-brain framework achieved consistent gains across all task categories: +19.0$\%$ in perception tasks, +20.0$\%$ in basic navigation, and up to +27.0$\%$ in advanced scenarios such as tunnel traversal. In the most complex case of obstacle avoidance, our method still outperformed the baseline by +19.0$\%$, underscoring the robustness and effectiveness of the proposed architecture in real-world deployment.

\begin{table*}[!htbp]
\centering
\caption{Performance comparison (\%)}
\begin{tabular}{lllccc}
\toprule
\textbf{Type} & \textbf{Level} & \textbf{Skill} & \textbf{Baseline \cite{ding2024quar} (Sim)} & \textbf{Ours (Real)} & \textbf{Improvement (\%)} \\
\midrule
Perception & Easy & Distinguish Letter  & 66\% & 51/60  & \textbf{+19\%} \\
Basic Navigation & Medium & Go to Object  & 60\% & 4/5  & \textbf{+20\%} \\
Advanced Capability & Hard & Go through Tunnel  & 53\% & 4/5 & \textbf{+27\%} \\
Advanced Capability & Hard & Go to Object and avoid obstacle  & 41\% & 3/5  & \textbf{+19\%} \\
\midrule
\multicolumn{3}{l}{\textbf{Training Data Required}} & 262K & \textbf{0} &  \\
\bottomrule
\end{tabular}
\end{table*}

This consistent performance gain across tasks of varying difficulty highlights the strength of our biologically inspired hierarchical control framework, which decouples high-level planning from low-level execution and thereby enables superior generalization and adaptability in real-world environments.

\subsection{Dual-brain effectiveness in field-realistic underwater conditions}

Compared with the idealized conditions of controlled laboratory tanks, real underwater environments exhibit significantly more complex and adverse physical characteristics. First, absorption and scattering by the water column drastically reduce ambient irradiance, leading to insufficient illumination and diminished contrast for imaging systems. Second, suspended particles such as silt, plankton, and organic detritus intensify back- and forward-scattering, producing hazy, color-shifted images with reduced fine detail. These impairments are further amplified in extreme settings like polar waters, deep-sea regions, or turbid coastal areas, posing critical challenges to vision-based underwater operations. 

To validate our framework's robustness under realistic conditions, we conducted experiments in a closed test tank with simulated marine degradation. Illumination was progressively reduced to mimic underwater luminance attenuation, while diatomaceous earth (5–200 µm diameter) was injected to raise turbidity from 0.5 NTU to 18 NTU (Fig.~\ref{fig:Task_Study}(A)-(D)). This created visual conditions consistent with natural turbid waters, reducing image contrast by ~65\% at 1.5 m range. 

We evaluated the dual-brain model (DBM) and single-brain end-to-end model (SBM) under two regimes: 1) standard laboratory conditions (normal lighting, low turbidity), and 2) simulated real-world conditions (low lighting + high turbidity). Results showed: 

\begin{itemize}
\item \textbf{Laboratory Conditions}: Both models completed tasks with high success rates, demonstrating comparable performance when visual inputs were ideal.
\end{itemize}

\begin{itemize}
\item \textbf{Simulated Real-World Conditions}: 
\begin{enumerate}
    \item \textbf{DBM Adaptation and Contextual Closure}: When visual degradation severely impaired target visibility (Fig.~\ref{fig:Task_Study}(C)), the DBM demonstrated the core strength of its hierarchical architecture. The cloud brain initially generated a high-level plan comprising four sub-tasks (``Rightward Alignment'', ``Medium-speed Forward'', ``Fine Pose Adjustment'', ``Approach and Maintain a Safe Distance''). Crucially, \textbf{the cerebellum (local executor) did not mechanically execute all four steps, but instead exhibited autonomous contextual assessment and task closure based on real-time visual feedback}. After completing the first two steps, the cerebellum entered the third sub-task, ``Fine Pose Adjustment'', and executed two low-speed maneuvers (a right turn followed by a left turn) to achieve precise relative positioning. Upon confirming that the robot had safely approached and stabilized at an appropriate standoff distance from the target, \textbf{the cerebellum autonomously judged that the safety criterion was satisfied and terminated further motion — deliberately omitting explicit execution of the fourth sub-task, since the objective of ``maintaining safe distance'' was already fulfilled through positional stabilization}. This behavior highlights the reactive layer’s capacity for dynamic, perception-driven decision-making: it not only follows the plan but also interprets task semantics and environmental context to determine when objectives are met, enabling efficient and safe mission closure.

    \item \textbf{SBM Failure Mode}: In contrast, the SBM lacked this adaptive capability. Upon losing the target (Fig.~\ref{fig:Task_Study}(b)), it continued executing its last valid command (``forward'') without replanning or distance regulation, eventually overshooting the goal area and violating safety margins. This failure stemmed from its inability to perform long-term state evaluation or adjust behavior based on degraded sensory input — particularly the lack of explicit safety-aware termination logic.
\end{enumerate}
\end{itemize}

These results demonstrate that the dual-brain architecture's separation of strategic planning and reactive execution provides critical resilience in optically challenging environments. The DBM's ability to initiate adaptive behaviors through explicit sub-task decomposition coupled with low-level autonomous adjudication (Fig.~\ref{fig:Task_Study}(A)→ (D)) contrasts sharply with the SBM's rigid, myopic response, validating our framework's practical utility for real-world underwater applications.

\begin{figure*}[!htbp]
  \centering
  \includegraphics[width=0.78\textwidth]{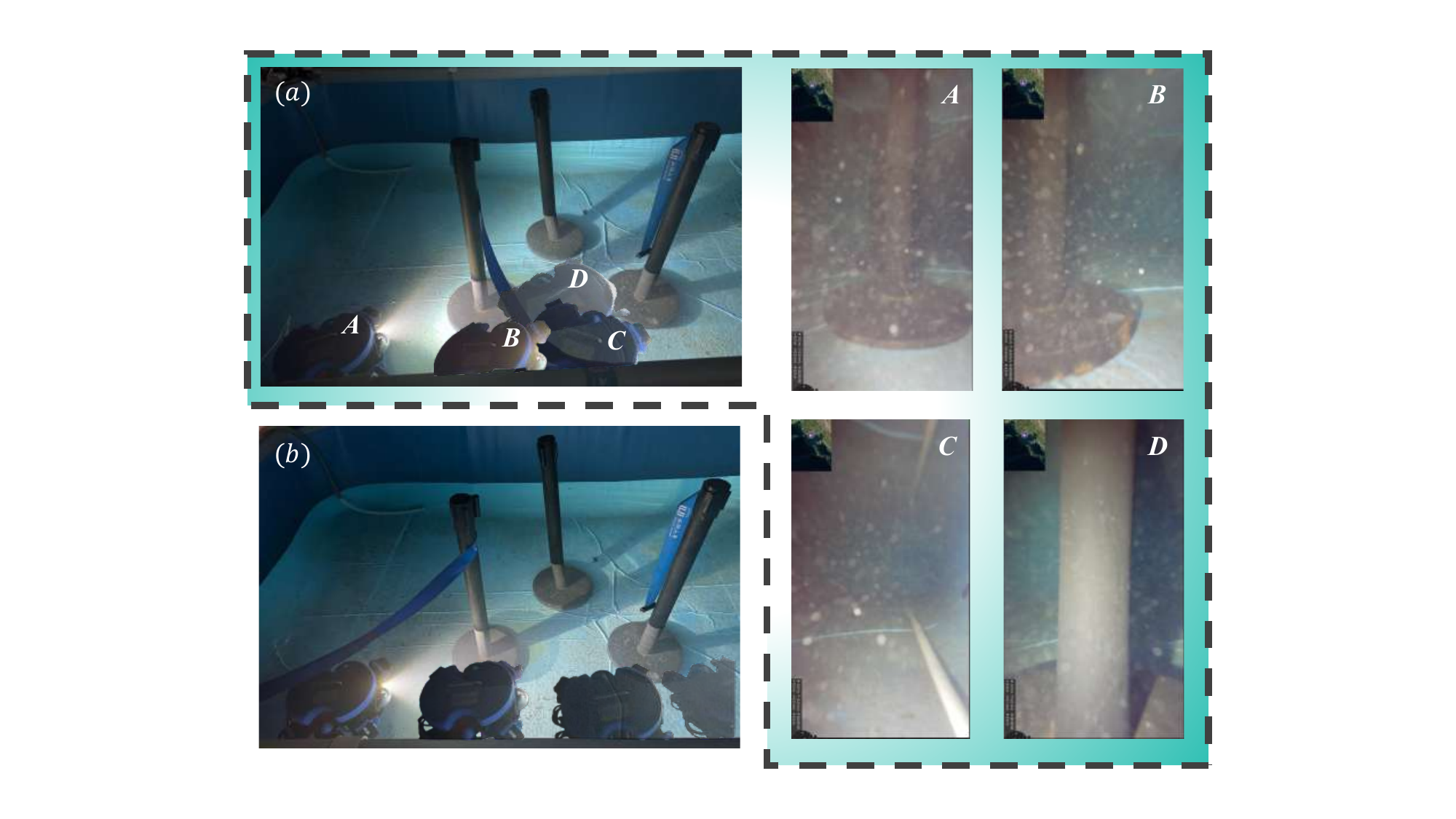} % Replace with the actual filename of your image
    \caption{Robustness comparison under simulated marine degradation: DBM vs. SBM. \textbf{(a)} and \textbf{(b)} illustrate the trajectory paths of the Dual-Brain Model (DBM) and Single-Brain Model (SBM), respectively, in degraded conditions. \textbf{(A)--(D)} show first-person visual perspectives at key decision points during DBM operation.
}
  \label{fig:Task_Study}
\end{figure*}

\section{Conclusion}

This work introduced \textbf{UnderwaterVLA} — the first Vision–Language–Action (VLA) framework specifically designed for autonomous underwater vehicles (AUVs), systematically overcoming the longstanding barriers that have hindered VLA adoption in marine robotics: the absence of an underwater-suited control hierarchy, dependence on costly demonstration data, and lack of real-time hydrodynamic compensation.

By introducing a biologically inspired \textit{dual-brain architecture}, we successfully decoupled high-level mission reasoning from low-level reactive control, enabling robust and interpretable autonomy under severe bandwidth and sensing constraints. Our \textit{zero-data training paradigm}, built upon prompt engineering and physics-informed Model Predictive Control, bypasses the need for underwater demonstrations while achieving state-of-the-art performance in real-world navigation and obstacle avoidance.

Experimental results demonstrate that UnderwaterVLA consistently outperforms conventional end-to-end and simulation-trained baselines — with +19\% to +27\% gains in perception, navigation, and complex instruction-following tasks. Crucially, it maintains graceful degradation and recovery under turbid, low-light conditions where traditional models fail catastrophically.

Looking ahead, we plan to extend UnderwaterVLA toward multi-agent coordination, long-duration missions, and extreme-depth operations. Future work will integrate real-time multimodal fusion, expand language grounding for domain-specific scientific commands (e.g., “survey hydrothermal vent at 2500 m”), and validate performance in full-scale open-ocean deployments. UnderwaterVLA thus lays the groundwork for a new class of linguistically grounded, physically aware, and data-efficient marine robots — transitioning embodied AI from simulated or lab-constrained environments into the dynamic, uncertain, and communication-starved reality of the deep sea.

\section*{Acknowledgment}

This work was supported by the National Key R\&D Program of China (Grant No. 2023YFC2806500) and the Zhejiang Province Key R\&D Program (Grant No. 2023C03133). Field test platforms were provided by DeepSeaTech Laboratory.

The authors acknowledge the use of Qwen (Qianwen), a large-scale generative AI model developed by Alibaba Cloud, to assist in:
(1) language polishing and stylistic refinement of the manuscript text;
(2) generating the initial code framework and assisting in debugging for data processing.

All technical content, experimental design, results, and conclusions were developed and validated solely by the authors.

%---------- IEEE 参考文献 ----------
\bibliographystyle{IEEEtran}
% 把你的 refs.bib 放到同目录，或者用 \input{refs.bib}
\bibliography{Springer_Nature_LaTeX_Template__2_/sn-bibliography}          % <--- 这里换成你自己的 bib 文件名（不含 .bib）

\end{document}